\documentclass[journal,twoside,web]{ieeecolor2}
\usepackage{generic}
\usepackage{cite}
\usepackage{amsmath,amssymb,amsfonts}
\usepackage{algorithmic}
\usepackage{graphicx}
\usepackage{textcomp}
\usepackage{multirow}
\usepackage{subfigure}

\def\BibTeX{{\rm B\kern-.05em{\sc i\kern-.025em b}\kern-.08em
    T\kern-.1667em\lower.7ex\hbox{E}\kern-.125emX}}
\begin{document}
\bibliographystyle{IEEEtran}

\title{Assistive Control of Knee Exoskeletons for Human Walking on Granular Terrains}
\author{Chunchu Zhu, Xunjie Chen, and Jingang Yi
\thanks{Manuscript submitted 15 December 2024; This work was supported in part by the US NSF under award CMMI-2222880. (Corresponding author: Jingang Yi.)}
\thanks{Chunchu Zhu, Xunjie Chen, and Jingang Yi are with the Department of Mechanical and Aerospace Engineering, Rutgers University, Piscataway, NJ 08854 USA (email: chunchu.zhu@rutgers.edu, xunjie.chen@rutgers.edu, jgyi@rutgers.edu).}}

\maketitle

\begin{abstract}
Human walkers encounter diverse terrains, such as sand and solid ground, which result in varied gait locomotion and energy costs. This study aims to design and validate a stiffness-based model predictive control approach for knee exoskeleton assistance with walking on sand. A comparative analysis of gait and locomotion on sand versus solid ground is conducted. A machine learning-based estimation scheme is developed to predict ground reaction forces in real time. These predictions, combined with human joint torque estimates, are used to design a knee exoskeleton controller that employs a model predictive stiffness control strategy. The experiments demonstrate significant differences in lower limb kinematics and kinetics. Reductions in major muscle activation and metabolic cost during assisted walking on sand are also observed with $15\%$ and $3.7\%$, respectively. The proposed knee exoskeleton control framework improves walking efficiency on granular terrains by reducing both muscular and metabolic demands.
\end{abstract}


\section{Introduction}
\label{sec:introduction}

Human often traverse real-world environments with a variety of terrains that can disrupt steady gait and require additional effort~\cite{kowalsky2021human}. While significant research has focused on walking on solid ground~\cite{camargo2021comprehensive}, real-world locomotion often occurs on granular terrains like sand, and this poses unique challenges due to the terrain's deformable and shifting properties~\cite{dewolf2019off}. Granular terrains significantly alter the biomechanic characteristics of walking gait, including joint-level kinematics, kinetics, and energy expenditure. Studies have shown that walking on sand requires large ankle plantarflexion in stance phase and increased knee and hip flexion during leg swing~\cite{van2017effect}. Substantial changes on walking kinematics and greater postural instability were observed for walking on sand~\cite{pinnington2005kinematic}. Similar studies of walking on compliant substrates demonstrated elevated activity and great mechanical work through large excursion and maximum flexion in hip and knee joints~\cite{grant2022does}. It has also been shown that walking on sand increases energy and metabolic cost~\cite{zamparo1992energy,grant2022does}, shifts center-of-pressure trajectories~\cite{xu2015influence}, and induces greater mechanical work~\cite{lejeune1998mechanics}. The work in~\cite{Zhu2024SandJBME} offered ground reaction forces and gait datasets with biomechanics of walking on solid ground and sand. Recent investigations into foot-substrate interactions highlight the influence of foot shape and speed on energy expenditure~\cite{chen2023energy,grant2024human}. These studies underscore the importance of terrain-specific insights to understand and optimize human walking locomotion.

Although walking on sand has been explored, limited studies are reported for using wearable assistive devices such as exoskeletons for enhancing the walking stability and reducing metabolic expenditure. Recent advancements in wearable exoskeletons and prosthesis development have shown promising capability in restoring ambulatory function~\cite{gehlhar2023review}, aiding postural adjustment, reducing muscle load~\cite{Zhu2021Design}, and decreasing metabolic cost during walking locomotion~\cite{panizzolo2019metabolic}. Knee exoskeletons have emerged as effective tools for enhancing mobility and stability, particularly in individuals with gait impairments or during challenging locomotion tasks~\cite{lee2021biomechanical}. Recent studies show that knee exoskeleton-assisted walking improves joint kinematics and enhance gait efficiency and stability by modulating neural activation and reducing muscle effort in individuals with impaired motor control~\cite{afzal2022evaluation,lee2023reducing}. Various control methods have been developed, including quasi-stiffness approaches that match target profiles~\cite{huang2022modeling} and reinforcement learning for impedance control~\cite{wu2022reinforcement}. Novel impedance modulation strategies have been proposed for lower-limb exoskeletons to assist sit-to-stand movements while preserving human control~\cite{huo2021impedance}, and for soft exoskeletons to aid push-off on various terrains~\cite{li2022human}. Trajectory tracking controls have been applied to lower limb exoskeletons to personalize and stabilize user gait movements with adaptive strategies~\cite{sharifi2022autonomous,shushtari2021online}. Model predictive control (MPC) was developed for compliant device behavior~\cite{bednarczyk2020model} and on-the-fly transitions between assistance modes~\cite{HunteAIM2020,jammeli2021assistive}.

\setcounter{figure}{1}
\begin{figure*}[ht]
		\hspace{-2mm}
		\subfigure[]{		
			\includegraphics[width = 2.44in]{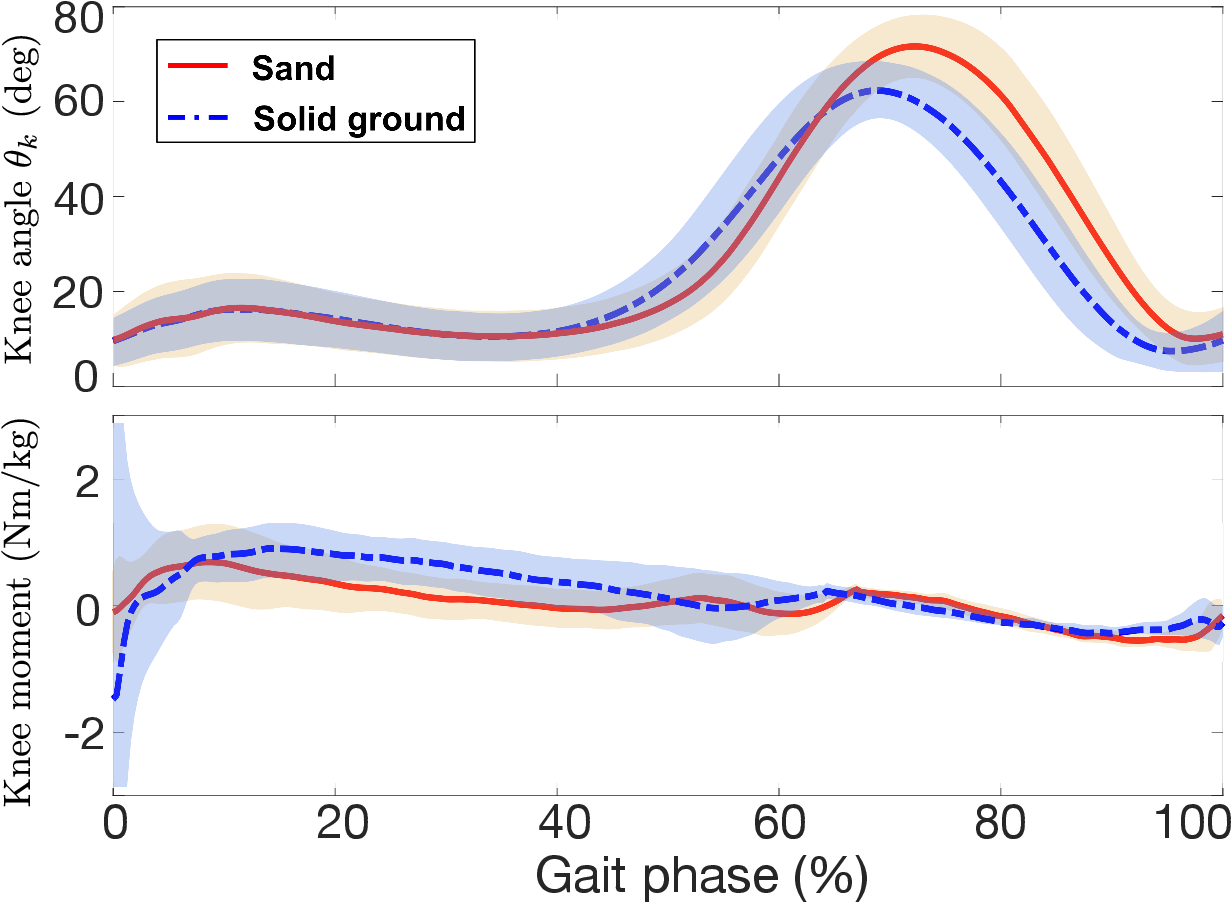}
			\label{fig:AngleAndMoment}}
		\hspace{-4mm}
		\subfigure[]{		
			\includegraphics[width = 2.4in]{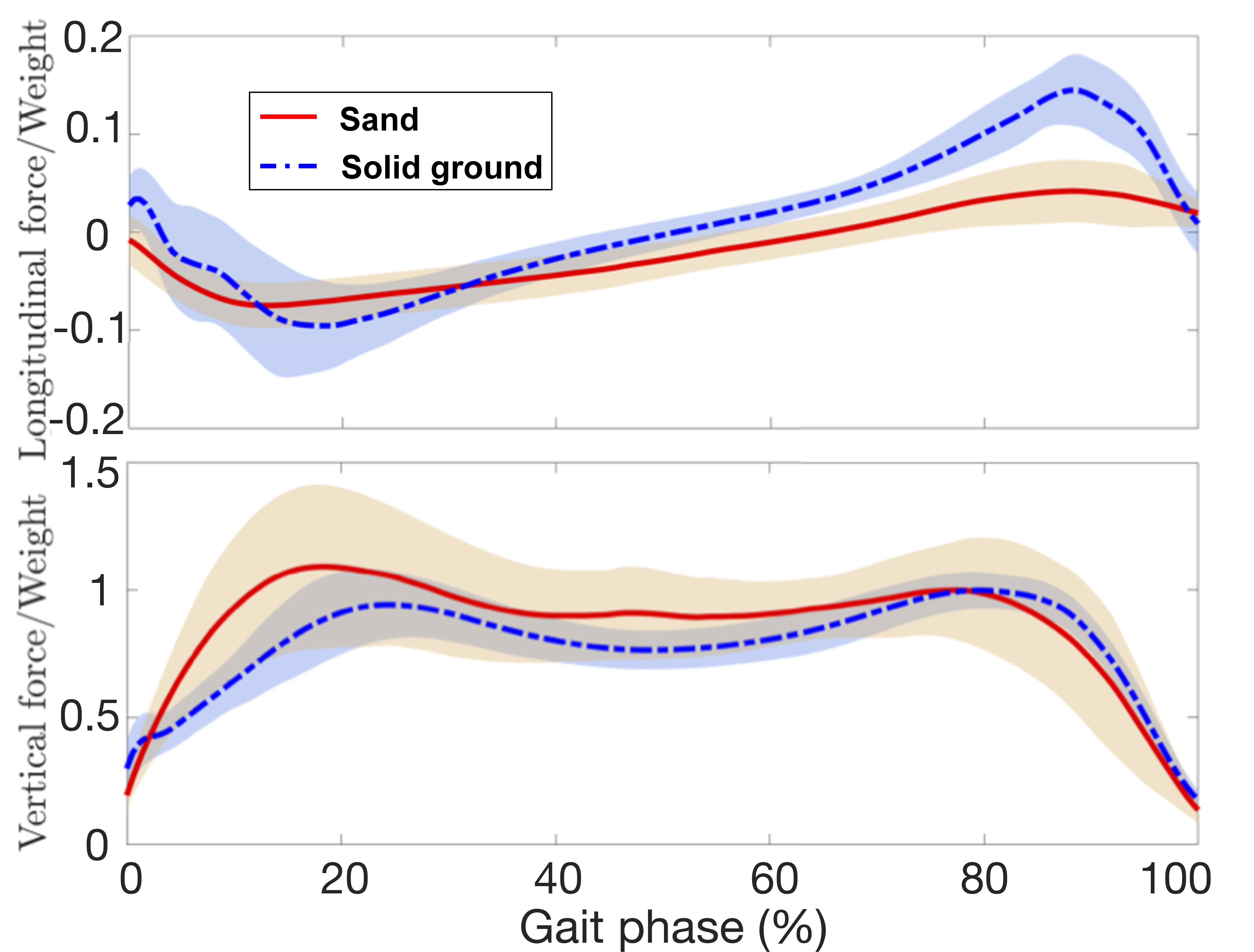}
			\label{fig:GRF}}
		\hspace{-4mm}
		\subfigure[]{		
			\includegraphics[width = 2.3in]{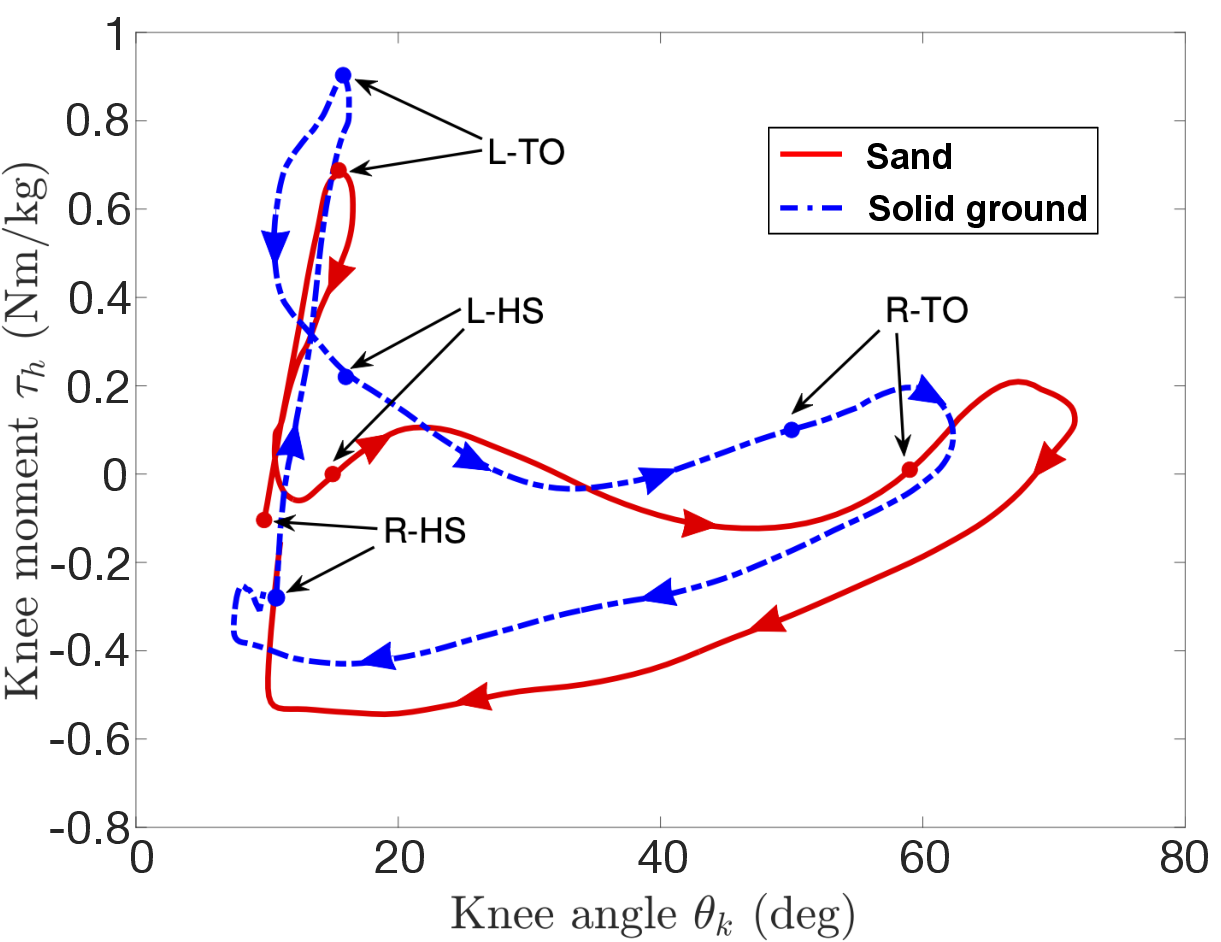}
			\label{fig:stiffness}}
			\vspace{-3mm}
	\caption{(a) Knee angle $\theta_k$ (top) and torque $\tau_h$ (bottom) profiles as function of gait phase $s$ of human walking on solid ground and sand. (b) The ground reaction forces on sand and solid ground: Longitudinal force $F_x$ (top) and vertical forces $F_z$ (bottom). (c) Knee joint stiffness curve during the walking gait. The arrows indicate the gait phase progressing direction. Heel strike (HS) and toe-off (TO) stand for heel-strike and toe-off events, respectively, and R and L for right and left legs, respectively.}
	\label{fig:biomechanics}
	\vspace{-5mm}
\end{figure*}

All aforementioned exoskeleton applications were primarily designed for human walking on solid ground, and the developed controllers cannot be directly applied to granular terrains. Moreover, the effect of knee exoskeleton control on joint moments among the ankle, knee, and hip joints on sand remains largely unexplored. To bridge this gap, we hypothesize that: (1) knee robotic assistance can reduce muscle activation and energy expenditure during walking on sand; and (2) knee exoskeleton control would alter the distribution of joint moments and joint angles among the ankle, knee, and hip joints for adapted walking efficiency. To test these hypotheses, we focus on designing and evaluating a knee exoskeleton for walking assistance on sandy terrains, aiming to enhance ambulation ability while reducing metabolic cost. We propose a model predictive stiffness control design of human-exoskeletons to provide assistive knee joint torque and regulate walking gait on sand. A stiffness-based computation of desired knee torque is first designed using the observations of walking locomotion on sand and solid ground. A neural network is used to model human foot-terrain interactions and compute resultant ground reaction forces (GRFs) from measurements by wearable inertial measurement units (IMUs). Both desired knee torque and external GRFs are treated as inputs of the human-exoskeleton dynamic model. Finally, an MPC framework is formulated to regulate assistive torque throughout the gait phase. Both indoor terrain and outdoor experiments are conducted to validate the controller performance.

The contributions of this study are twofold: (1) a detailed analysis of joint-level biomechanics during walking on solid ground and sand, focusing on GRFs, joint torques, and energy cost; and (2) the design and validation of an adaptive exoskeleton control framework for granular terrains, emphasizing biomechanical insights and real-time adaptability. By addressing these biomechanical challenges, this study not only extends the applicability of wearable assistive devices to diverse environments but also provides novel insights into the dynamics of human-exoskeleton interaction on yielding terrains. The proposed stiffness-based MPC framework improves energy efficiency and gait stability on compliant surfaces, offering significant implications for assistive device design and rehabilitation. By examining the distribution of joint moments and angles across the hip, knee, and ankle joints, we highlight the role of knee assistance in redistributing mechanical effort and reducing metabolic cost. 



\section{Exoskeleton-Assisted Human Walking on Granular Terrain}
\label{sec:SystemDesign}

\subsection{Gait and Locomotion on Granular Terrains}

Compared to solid ground, human walking on sand shows different gait and locomotion characteristics. Fig.~\ref{fig:photo} shows the human subject with wearable IMUs and bilateral knee exoskeletons that are used in experiments. Fig.~\ref{fig:Model} illustrates the kinematics and modeling configuration of the human-exoskeleton interactions on sand. A two-link limb model is used with joint angle $\boldsymbol{q} = [\theta_{t} \; \theta_{k}]^T \in \mathbb{R}^2$, where $\theta_{t}$ and $\theta_k$ represent the thigh and knee angle, respectively. The longitudinal and vertical GRFs are denoted as $F_x$ and $F_z$, respectively. The walking gait progression, namely, the joint angles and moments, was normalized by a gait phase variable~$s$ (between 0 and 1) according to the heel strike events.

\setcounter{figure}{0}
\begin{figure}[h!]
	\vspace{-1mm}
		\subfigure[]{\includegraphics[width = 1.6in]{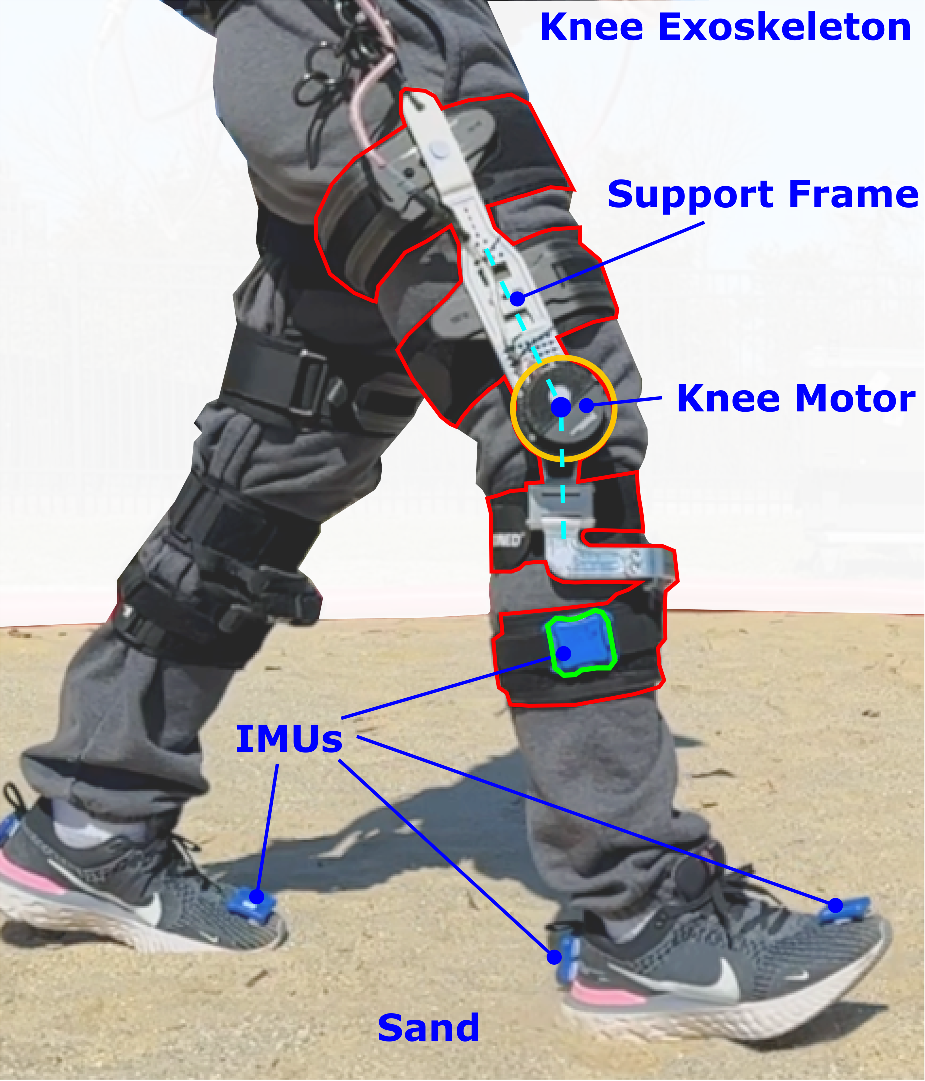}
		\label{fig:photo}}
	\hspace{-2mm}
		\subfigure[]{\includegraphics[width = 1.7in]{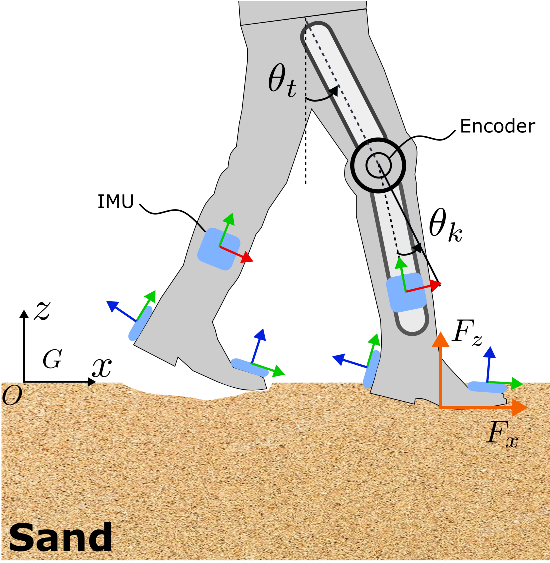}
		\label{fig:Model}}
		\caption{(a) Subject with IMUs and bilateral exoskeletons. (b)~Schematic of the human lower limb and human-exoskeleton interaction during the swing (left leg) and stance (right leg) phases on sand.}
		\label{humansetup}
		\vspace{-1mm}
\end{figure}

Using experimental data from~\cite{Zhu2024SandJBME}, we compare biomechanical features of walking on sand versus solid ground. Fig.~\ref{fig:AngleAndMoment} shows the knee angle ($\theta_k$) and joint moment ($\tau_h$) over the gait phase $s$ for walking on both terrains. Fig.~\ref{fig:GRF} compares the GRFs from experiments with $20$ human subjects. The comparative analysis reveals increased angular displacement and reduced moment amplitude on sand. The longitudinal force $F_x$ and vertical force $F_z$ clearly exhibit distinct patterns when walking on sand compared to solid ground. While the hip and ankle contribute to propulsion and mechanical work during walking on sand, the knee joint stabilizes gait on granular terrain, therefore enhancing stability and gait efficiency. Fig.~\ref{fig:stiffness} illustrates the relationship between $\tau_h$ and $\theta_k$ over the gait phase on both terrains, where the slope indicates joint stiffness. Differences in knee stiffness are minimal from right heel strike (R-HS) to left toe-off (L-TO) but become significant during the swing phase, i.e., R-TO$\rightarrow$R-HS. The stiffness curve demonstrates a notable shift while maintaining its linear characteristic. A stiffness-based model is used to estimate real-time knee joint torques.

Similar to~\cite{huang2022modeling}, a Sigmoid function is considered to capture a smooth transition between stance and swing phases of the gait cycle, that is,
\begin{equation}
    S(\boldsymbol{\theta}_k)=\frac{1}{1+e^{-a f(\boldsymbol{\theta}_k)}},
   \label{eq0}
\end{equation}
where $\boldsymbol{\theta}_k=[\theta_{kr} \;\, \theta_{kl}]^T$, $\theta_{kl}$ and $\theta_{kr}$ are the left and right knee joint angles, respectively. Function $f(\boldsymbol{\theta}_k)=(\theta_{kr}-\theta_{kl})-b$ represents an estimated optimal hyperplane that divides the gait cycle into two phases. The parameters $a$ and $b$ are optimized based on collected human subject data. Using~\eqref{eq0}, the estimated human knee moment is calculated by
\begin{equation}
\hat{\tau}_{h}=\left[1 - S(\boldsymbol{\theta}_k)\right] k_{st}(\theta_{ki}-\theta^0_{st})+S(\boldsymbol{\theta}_k) k_{sw}(\theta_{ki}-\theta^0_{sw})
\label{esttorq}
\end{equation}
with $i=l,r$ for left and right legs, respectively. In~\eqref{esttorq}, $k_{st}$ ($k_{sw}$) is the joint stiffness in the stance (swing) phase, and $\theta^0_{st}$ ($\theta^0_{sw}$) is the equilibrium angle for stance (swing) phase. Model parameters such as $k_{st}$, $k_{sw}$, $k^0_{st}$, $k^0_{sw}$, $a$, and $b$ are estimated by minimizing the sum of squared errors between the estimated and actual knee moments. The actual knee torque is calculated from the ground-truth GRF measurements.

\vspace{-1mm}
\subsection{Learning-based GRF Estimation}
\label{subsec:learningGRF}

Estimating GRFs is essential for real-time exoskeleton control on sand, as direct measurements are impractical. A machine learning-based model predicts GRFs using temporal and spatial features from IMU data, offering a non-invasive, real-time solution by capturing motion and force dynamics without ground contact. Fig.~\ref{fig:network} illustrates a dual-pathway machine learning model for real-time GRF prediction. Input data are from IMUs mounted on the stance leg's shank, heel, and toe; see Fig.~\ref{fig:Model}. Similar to~\cite{YigitTASE2022}, only the accelerations in the sagittal plane ($x$- and $z$-axis) and the frontal-axis gyroscopic measurements ($y$-axis) from each IMU are used as the input for the machine learning model, i.e., $\boldsymbol{X}_t=[a_x(t) \; a_z(t) \; \omega_y(t)]^T$ from all three IMUs at time $t$. 

\setcounter{figure}{2}
\begin{figure}[ht!]
	\centering
		\includegraphics[width = 3.35in]{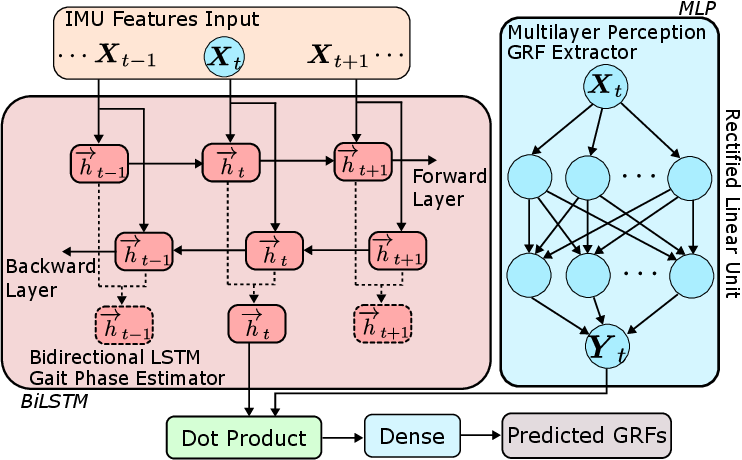}
		\vspace{-1mm}
		\caption{Schematic of the machine learning-based dual-pathway GRFs prediction scheme.}
    \label{fig:network}
	\vspace{-2mm}
\end{figure}

The first pathway employs a bidirectional long short-term memory (Bi-LSTM) layer with 64 units to detect foot contact periods and gait phases. By processing data bidirectionally, the Bi-LSTM simultaneously learns from both past and future contexts. The hidden state output $\boldsymbol{h}_t$ from the Bi-LSTM captures the temporal dynamics essential for identifying these periods and phases. The second pathway utilizes a multi-layer perceptron (MLP) to estimate terrain conditions and continuous GRF magnitudes during the detected stance phases. $\boldsymbol{Y}_t $ in the figure represents the terrain conditions and GRF estimates, derived from an MLP comprising two dense layers (128 and 64 units) with rectified linear unit activation. This dual-pathway architecture is designed to capture the relationships between temporal movement sequences (e.g.,  $ \ldots,\boldsymbol{X}_{t-1},\boldsymbol{X}_{t},\boldsymbol{X}_{t+1},\ldots$) and their spatial configurations for accurate GRF prediction. A sliding window length of 64 ms was used to enable timely updates for real-time application.

Outputs from LSTM and MLP are combined via dot product, emphasizing temporal-spatial feature interaction for a more accurate GRF estimation. A final dense layer with linear activation predicts GRFs. The model is trained using mean squared error loss and Adam optimizer, with early stopping and learning rate scheduling for regularization and optimization. This architecture is hypothesized to outperform traditional single-pathway models~\cite{alcantara2022predicting} by explicitly modeling both temporal progression and spatial configurations.

\subsection{Stiffness-Based Exoskeleton Control}
\label{sec:Control}

\begin{figure}[h!]
	\centering
	\includegraphics[width = 3.4in]{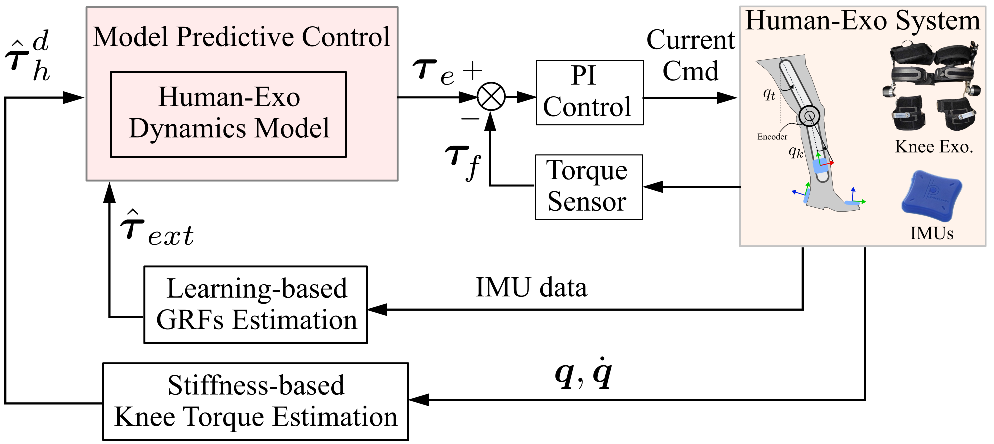}
	\caption{The schematic of the overall exoskeleton control design.}
	\vspace{-2mm}
	\label{fig:control}
\end{figure}

Fig.~\ref{fig:control} illustrates the overall knee exoskeleton control design. The control scheme is built on real-time estimation of human torque $\hat{\boldsymbol{\tau}}_h$ from the stiffness-based model and external torque $\hat{\boldsymbol{\tau}}_{ext}$ via a machine learning model of the GRFs. IMUs and encoders provide the gait phase variable $s$ and lower-limb joint kinematics. An MPC scheme is formulated to design the controlled exoskeleton torque, denoted by $\boldsymbol{\tau}_e$. The lower-level controller of the exoskeletons takes the MPC outcome to drive the human-exoskeleton system to maintain desired joint angle and torque profiles on different terrains. The exoskeleton lower-level control is designed as a proportional-integral (PI) controller and the feedback torque signal is from the motor current sensor. 

While other control strategies might achieve joint angle trajectory tracking, MPC’s predictive capabilities are essential for managing the non-linear walking dynamics on sand. By optimizing control inputs over a horizon, MPC demonstrates particularly effectively maintaining balance on deformable surfaces. The objective of the MPC has two key aspects: follow optimal joint angle trajectories and minimize the effort required by both the human and the exoskeleton. The cost function was chosen to ensure smooth torque assistance and stability, thus minimizing the physical load on the user.


To simplify the dynamics while maintaining accuracy, we assume bilateral symmetry and model point contact at the ankle joint. This allows us to focus on the stance leg dynamics with the exoskeleton in the sagittal plane; see Fig.~\ref{fig:Model}. The two-link robot dynamics is captured by 
\begin{equation}
    \boldsymbol M(\boldsymbol q) \ddot{\boldsymbol q} + \boldsymbol C(\boldsymbol q, \dot{\boldsymbol q}) \dot{\boldsymbol q} + \boldsymbol G(\boldsymbol q) = \boldsymbol \tau_e + \boldsymbol \tau_h + \boldsymbol \tau_{\text{ext}},
    \label{eq:model}
\end{equation}
{where $\boldsymbol{M}(\boldsymbol{q})$, $\boldsymbol{C}(\boldsymbol{q},\dot{\boldsymbol{q}})$, and $\boldsymbol{G}(\boldsymbol{q})$ are the subject-dependent inertia, Coriolis force, and gravitational force matrices, respectively.} Human applied torque is $\boldsymbol{\tau}_h=[0 \; \tau_h]^T$, exoskeleton torque $\boldsymbol{\tau}_e=[0 \; \tau_e]^T$, and $\boldsymbol{\tau}_{\text{ext}}= \boldsymbol{J}^{T} \boldsymbol{F}_{\text{ext}}$, with $\boldsymbol{F}_{\text{ext}}=[F_x \; F_z]^T$ as the GRFs and $\boldsymbol{J} \in \mathbb{R}^{2\times 2}$ as the Jacobian matrix. Although walking on sand does increase the foot contact area compared to solid ground, we approximate the foot-terrain contact as a point because of its relatively small range. This simplification allows for efficient real-time implementation without significantly compromising control accuracy.

Using~\eqref{esttorq} and the GRF estimation, we obtain the estimates $\hat{\boldsymbol{\tau}}_h$ and $\hat{\boldsymbol{\tau}}_{\text{ext}}=\boldsymbol{J}^{T} \hat{\boldsymbol{F}}_{\text{ext}}$ for $\boldsymbol{\tau}_h$ and $\boldsymbol{\tau}_e$ in~\eqref{eq:model}, respectively. Using these estimated torques, we rewrite~(\ref{eq:model}) as
\begin{equation}
\boldsymbol{M}(\boldsymbol q) \ddot{\boldsymbol q} + \boldsymbol C(\boldsymbol q, \dot{\boldsymbol q}) \dot{\boldsymbol q} + \boldsymbol G(\boldsymbol q) = \boldsymbol \tau_e + \hat{\boldsymbol{\tau}}_h + \hat{\boldsymbol{\tau}}_{\text{ext}}+\Delta \boldsymbol{\tau},
\label{eq:estimatedModel2}
\end{equation}
where $\Delta \boldsymbol{\tau}= (\boldsymbol{\tau}_h-\hat{\boldsymbol{\tau}}_h) + (\boldsymbol{\tau}_{\text{ext}}-\hat{\boldsymbol{\tau}}_{\text{ext}})$ is the torque estimation error. The desired assistive torque provided by the exoskeleton is denoted as $\boldsymbol{\tau}^d_e$ and is considered as 
\begin{equation}
	\boldsymbol{\tau}_e^d = \alpha \hat{\boldsymbol{\tau}}_h,
	\label{desiredtorq}
\end{equation}
where $\alpha >0$ is a scaling factor for the level of assistance.

For presentation convenience, we use discrete-time form of the systems dynamics given by~\eqref{eq:model} and drop the dependence of matrices $\boldsymbol{M}$, $\boldsymbol{C}$ and $\boldsymbol{G}$ on $\boldsymbol{q}$ and $\dot{\boldsymbol{q}}$. The MPC is formulated as follows,
\begin{subequations}
	\begin{align}
		\hspace{-4mm}\min_{\boldsymbol{\tau}_E} \sum_{i=m}^{m+H} [w_1(\Delta \dot{\boldsymbol{q}}(i))^2&+ w_2 (\Delta \boldsymbol{\tau}_e(i))^2 + w_3 (\Delta \boldsymbol{\tau}_e^d(i))^2],  \label{mpca}\\
		\text{subj. to:}~ \boldsymbol{q}(i+1)=&\boldsymbol{q}(i)+\Delta t \dot{\boldsymbol{q}}(i), \label{mpcb} \\ 
				\dot{\boldsymbol{q}}(i+1)=&\dot{\boldsymbol{q}}(i)+\Delta t\boldsymbol{M}^{-1}_i\Bigl[\boldsymbol{\tau}_e(i)+\hat{\boldsymbol{\tau}}_h(i)+ \nonumber \\ 
				&\hat{\boldsymbol{\tau}}_{\text{ext}}(i)-\boldsymbol{C}_i\dot{\boldsymbol{q}}(i)-\boldsymbol{G}_i\Bigr], \label{mpcc} \\
	\underline{q} \leq \|\boldsymbol{q}(i)\|&\leq \bar{q},\; \underline{\dot{q}} \leq \|\dot{\boldsymbol{q}}(i)\| \leq \bar{\dot{q}},\\
	\underline{\ddot{q}} \leq \|\ddot{\boldsymbol{q}}(i)\| &\leq \bar{\ddot{q}},\; \,i=m,\cdots,m+H, \label{mpce}
	\end{align}
\end{subequations}
where $H \in \mathbb{N}$ is the prediction horizon, $m\in \mathbb{N}$, $\Delta t$ is the time step, and $\underline{q}$ ($\bar{q}$), $\underline{\dot{q}}$ ($\bar{\dot{q}}$), and $\underline{\ddot{q}}$ ($\bar{\ddot{q}}$) are the minimal (maximal) bounds for the joint angle, velocity, and acceleration, respectively. $\boldsymbol{M}_i$, $\boldsymbol{C}_i$, and $\boldsymbol{G}_i$ are the corresponding matrices $\boldsymbol{M}$, $\boldsymbol{C}$, and $\boldsymbol{G}$ evaluated at $t=i \Delta t$, $i\in \mathbb{N}$. In~\eqref{mpca}, joint angle error $\Delta \dot{\boldsymbol{q}}(i) =\dot{\boldsymbol{q}}_d(i) - \dot{\boldsymbol{q}}(i)$, $\Delta \boldsymbol{\tau}_e(i) = \boldsymbol{\tau}_e(i+1) - \boldsymbol{\tau}_e(i)$, and $\Delta \boldsymbol{\tau}_e^d(i) = \boldsymbol{\tau}_e^d(i+1) - \boldsymbol{\tau}_e(i)$, $i=m,\cdots,m+H$. $w_1, w_2$, and $w_3$ are the respective weights. The design variable to the MPC problem is $\boldsymbol{\tau}_E=\{\boldsymbol{\tau}_e(m),\,\boldsymbol{\tau}_e(m+1),\,\cdots,\boldsymbol{\tau}_e(m+H)\}$ and the first element of the MPC solution $\boldsymbol{\tau}_e(m)$ is chosen to implement at the $m$th step.

Compared to the robot dynamics in~\eqref{eq:estimatedModel2}, the constraint in~\eqref{mpcc} does not include the torque estimation term $\Delta \boldsymbol{\tau}$ since we do not have the exact information about human torque $\boldsymbol{\tau}_h$ and foot-terrain interaction force $\boldsymbol{F}_{\text{ext}}$. However, we assume that the stiffness model and the machine learning-enabled IMU-based estimations are accurate such that $\Delta \boldsymbol{\tau}$ is small and negligible and therefore, \eqref{mpcc} approximates the actual robot dynamics.

\section{Experimental Design}
\label{sec:Exp}

\subsection{Experiment Setup and Protocols}

Eight able-bodied healthy subjects (6 males and 2 females, age: $24.8 \pm 2.7$, height: $168 \pm 7.8$~cm, weight: $65.5 \pm 9.5 $~kg) participated in experiments, with all eight joining the indoor trials and five of them also participating in the outdoor trials. All participants were self-reported to be in a good health condition. An informed consent form was signed by each subject and the experimental protocol was approved by the Institutional Review Board (IRB) at Rutgers University.
\begin{figure*}[ht!]
	\hspace{-2mm}
	\subfigure{
		\includegraphics[width =2.1in]{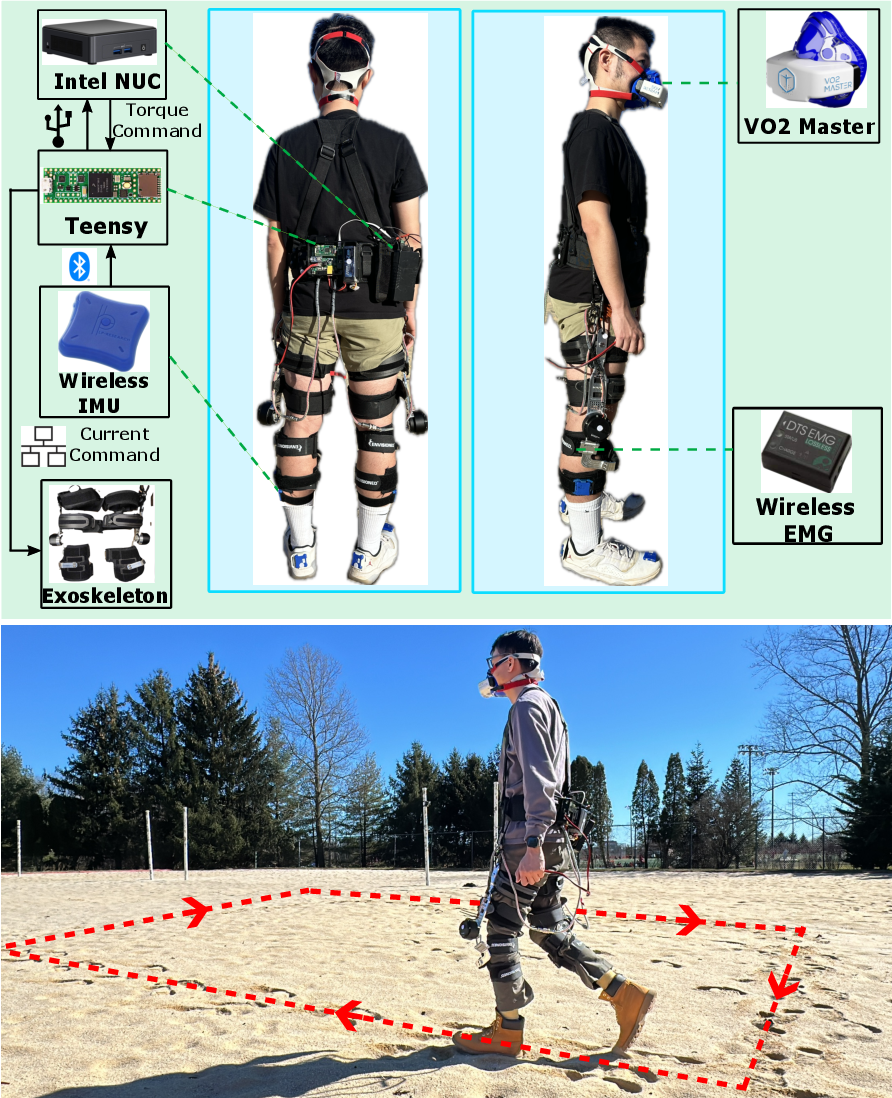}
		\label{fig:devices}}
	\hspace{-2.4mm}
			\subfigure{
			\includegraphics[width=4.8in]{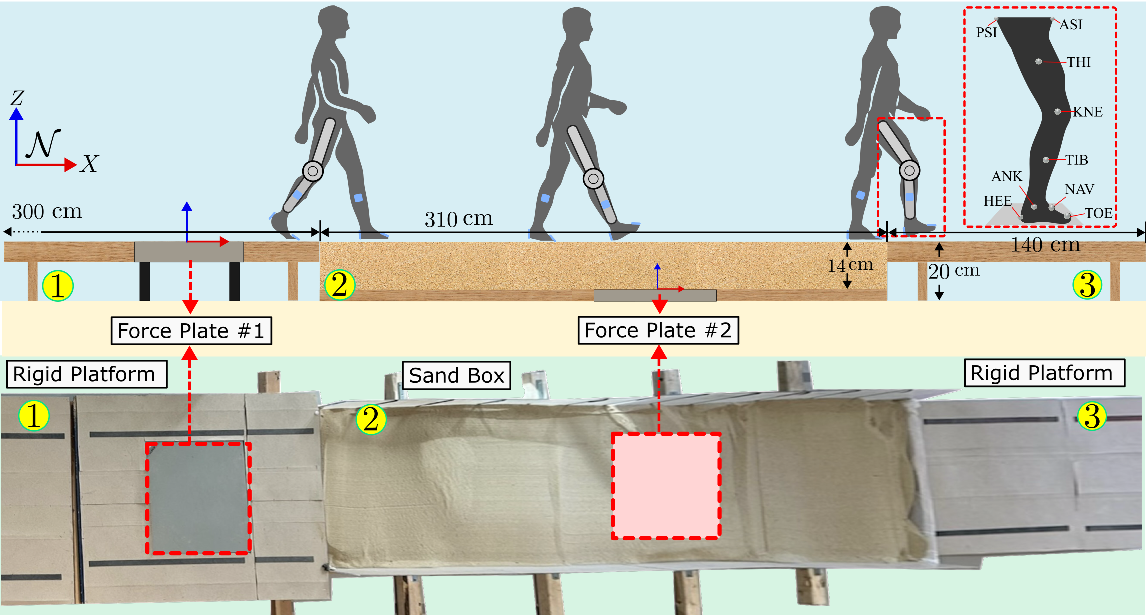}
			\label{fig:indoor}}
		\caption{(a) Experimental devices and the interconnection among various wearable sensing, exoskeletons and embedded systems (top) and outdoor experimental setup (bottom). (b) Indoor experimental setup with sand and solid ground platform with motion capture system (not shown in the figure).}
	\vspace{-3mm}
\end{figure*}

The top figure in Fig.~\ref{fig:devices} shows the wearable devices and their interconnections in experiments. Six IMUs (from LP-RESEARCH Inc.) were mounted on predefined locations on left and right shanks, heels and toes. We carefully checked and calibrated the sensors before each trial to minimize the potential for sensor movement during walking. These IMUs and the encoders of the bilateral knee exoskeletons were the primary sensing sources for exoskeleton control~\cite{ZhuRAL2023}. IMU data was wirelessly transmitted to a high-level micro-computer (NUC7i7DNK from Intel Corp.) for real-time estimation of human joint moments and GRFs. The MPC solver, implemented via customized Python scripts on the micro-computer, generated torque commands. These commands were relayed to a low-level microcontroller (from PJRC Teensy) for exoskeleton torque control via CAN bus. The control systems parameters were given as: $\theta_{st}^0 = 8.7$~deg, $\theta_{sw}^0 = 68.7$~deg, $k_{st} = 0.047$, $k_{sw} = 0.012$, $a = 0.19$, $b = 3.85$, $w_1 = 1$, $w_2 = 0.15$, $w_3 = 0.5$, $H = 50$, $\Delta t = 0.04$~s, and $\alpha = 0.3$.

Fig.~\ref{fig:indoor} shows the indoor experimental setup. Ground-truth lower-limb kinematics were captured using an optical motion capture system (10 Vicon Bonita cameras from Vicon Ltd.) A three-segment walkway ($7.5$~m long, $0.76$~m wide, and $0.14$~m high) was constructed for indoor experiments. Segment $1$ was built with reinforced plywood with embedded force plate $\#1$ (model ACG-O from AMT Inc.) to replicate a solid ground condition. Segment $2$ was a sand-filled box with buried force plate \#2 (from Bertec Corp.) 14 cm beneath the surface. GRF measurements were calibrated for sand layer thickness~\cite{Zhu2024SandJBME}. Segment $3$ was made of the same material as Segment $1$ to allow natural gait transition from sand. The three-segment walkway design ensures that the subjects employ the consistent walking strategy throughout the trial. For evaluation purposes, subjects were equipped with eight wireless EMG sensors (from Delsys Inc.) on the right leg, monitoring rectus femoris (RF), vastus lateralis (VL), vastus medialis (VM), biceps femoris (BF), semitendinosus (SEM), tibialis anterior (TA), lateral gastrocnemius (LG), and medial gastrocnemius (MG). The EMG data were sampled at $1500$~Hz, followed by notch filtering ($58$-$62$~Hz) to remove line noise and band-pass filtering ($30$-$500$~Hz with $4$th-order zero-phase Butter-worth filters). 

\subsection{Indoor/Outdoor Tests and Data Collection}
\label{exp:indoor}

For indoor experiments, subjects participated in four groups of experiments, including Group~A: no exoskeleton; Group~B: unpowered exoskeletons; Group~C: exoskeletons with the baseline controller from~\cite{huang2022modeling}; and Group~D: exoskeletons with the proposed MPC stiffness controller. The baseline controller was selected because it outperformed the most impedance controllers in minimizing muscle activation. The subjects were instructed to walk in their preferred manner in one direction starting at Segment $1$. For the first $5$ trials, the subject was required to walk on the walkway to get familiar with the sensing devices and the terrain conditions. After that, $5$ trials of each group were conducted, and paired sample t-tests were used for statistical comparisons across the groups, resulting in a total of 40 trials per group. Ten-minute breaks were provided between groups, and the sand surface was flattened before each trial. For analysis, the root-mean-square (RMS) and maximal values of each muscle's EMG signal were extracted for the steps in each group and terrain, averaged, and then normalized to the RMS or peak values in Group A.

Outdoor experiments (see bottom figure in Fig.~\ref{fig:devices}) involved subjects wearing the knee exoskeleton and VO2 Master metabolic measurement device (from Vernon Inc.). Participants were instructed to follow marked trajectories for approximately $10$ minutes ($500$ m) under the four groups (A-D), assigned randomly to eliminate order bias. Additionally, participants were instructed to maintain a consistent walking pace as their choices for comfort and capability. This approach ensured that the data collected would accurately reflect the impact of the exoskeleton under varied yet controlled conditions. Throughout the trials, the maximum rate of oxygen from the VO2 Master device was recorded for further analysis.

Key evaluation metrics included: (1) GRF prediction accuracy; (2) controller's influence on joint kinematics and kinetics; (3) impacts on lower-limb muscle activation; and (4) performance in reducing metabolic costs on different terrains. Results from indoor experiments were used to address the first three objectives, while outdoor experiments were conducted to evaluate metabolic costs over extended periods. To obtain accurate subject-specific dynamic models in~\eqref{eq:model}, a system identification is required for each subject. To do so, two experiment trials were conducted for each subject. The first trial identified the subjects' dynamics without the exoskeleton, while the second experiment for the human-exoskeleton system. Detailed experiment procedures and the optimization method can be found in~\cite{huo2021impedance}. 

For machine learning-based GRF estimation, we used a dataset from 20 subjects as detailed in~\cite{Zhu2024SandJBME}. The experiments followed the indoor setup in Fig.~\ref{fig:indoor}, with foot-force plate contacts on both solid ground and sand as discussed above. The IMU data were synchronized with motion capture and GRF recordings, down-sampled to $100$~Hz, and normalized by body weight to enhance model generalizability. GRF recordings~ were down-sampled to $100$~Hz to synchronize with the motion capture and IMU data. The GRF data were normalized to each subject's body weight to provide a dimensionless measure of the biomechanical impact of gaits.
\setcounter{figure}{7}
\begin{figure*}
		\hspace{-3mm}
		\subfigure[]{
		\includegraphics[width = 3.55in]{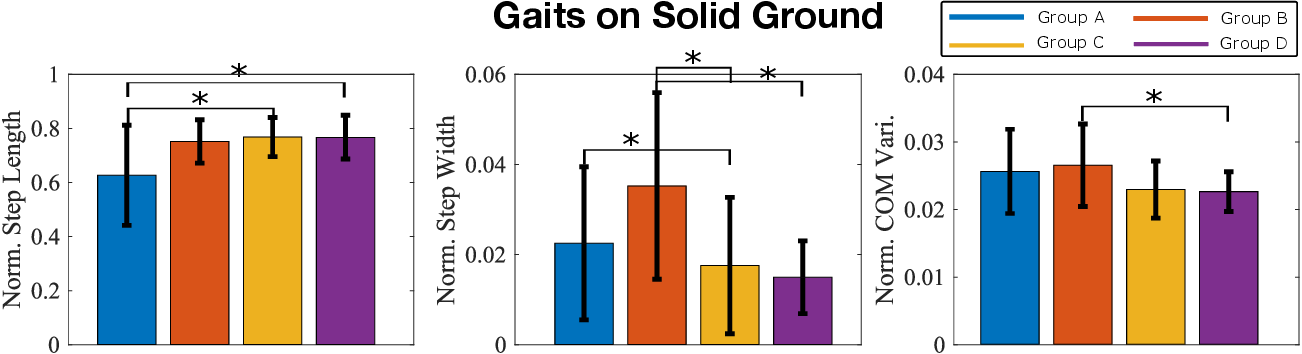}
		\label{fig:GaitOnSolid}}
		\hspace{-3.5mm}
		\subfigure[]{
		\includegraphics[width = 3.55in]{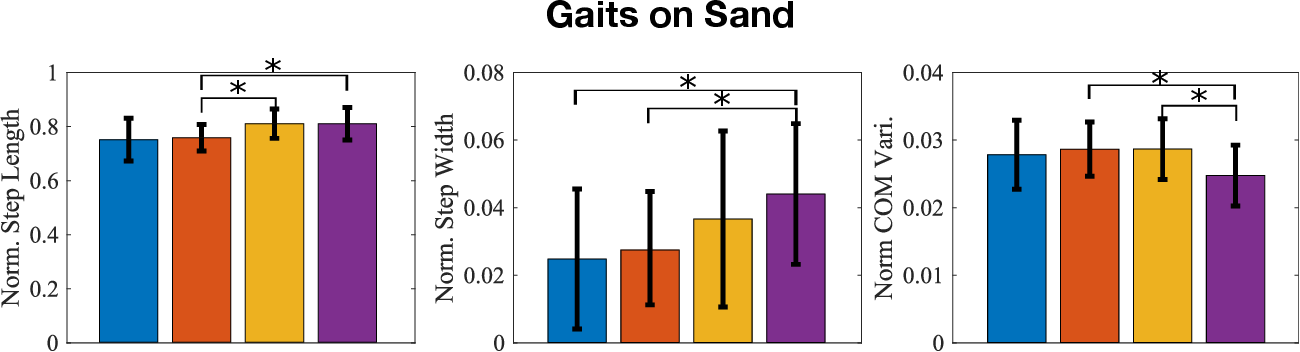}
		\label{fig:GaitOnSand}}
		\vspace{-3mm}
		\caption{Step parameters for participants walking on solid ground vs. sand: (a) solid ground and (b) sand. The comparison was among the stride length, stride width, and COM variation (all normalized with respect to subjects' heights). The label $*$ indicates a significant (p $< 0.05$) difference among groups.}
		\vspace{-3mm}
\end{figure*}

\section{Results}
\label{sec:results}

\subsection{Gait Kinematics and Kinetics}

The average self-selected walking speed across participants was approximately $1.02 \pm 0.08$~m/s. Fig.~\ref{fig:predGRF} shows the comparison between the actual and predicted GRFs (normalized by body weight) on sand in indoor experiments. The contact phase refers to the portion of the gait cycle during which the foot maintains contact with the ground. The prediction root mean square errors (RMSE) for normalized $F_x$ was $0.052\pm 0.015$ on solid ground and $0.12 \pm 0.027$ on sand. For normalized $F_z$, the RMSE was $0.11 \pm 0.018$ on solid ground and $0.11 \pm 0.021$ on sand. The observed higher variance in GRF predictions on sand was attributed to the deformable terrain. The lowest RMSE was observed for the $F_x$ component on solid ground. To statistically evaluate the differences in GRFs across terrains, a paired sample t-test was conducted and the analysis revealed a statistically significant difference ($p<0.05$) in the GRFs on solid ground and sand. This finding supports that the GRFs can be used for terrain classifications and terrain variability significantly influences the accuracy of GRF predictions.

\setcounter{figure}{5}
\begin{figure}[h!]
	\centering
		\includegraphics[width = 3.4in]{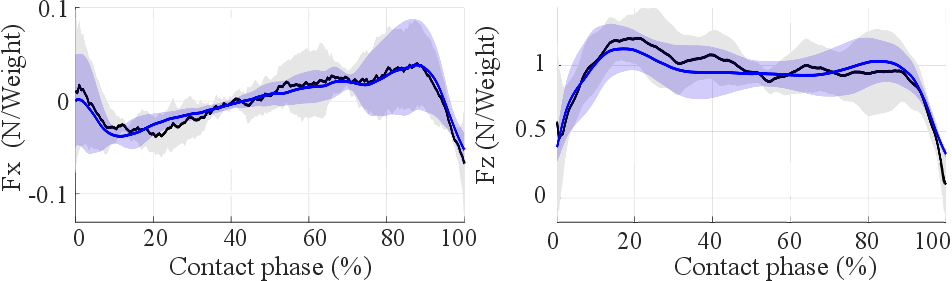}
	\vspace{-1mm}
	\caption{Comparisons of the actual and estimated GRFs in Group D on sand.}
	\label{fig:predGRF}
\end{figure}

Fig.~\ref{fig:torques} shows a clear comparison of average exoskeleton torque outputs between Groups C and D. The use of exoskeleton aims at reducing human effort rather than replacing natural knee torque entirely. Both the baseline controller and the MPC stiffness controller provided the more torque assistance during the stance phase than for the swing phase. Furthermore, the exoskeleton assisted human walking more significantly within the late stance phase (i.e., $35$--$65$\% of gait phase) on sand than on solid ground. It is also found that the MPC controller offered larger joint moment assistance on sand than that by the baseline controller during the swing phase.

\setcounter{figure}{6}
\begin{figure}[h!]
	\centering
	\includegraphics[width = 3.4in]{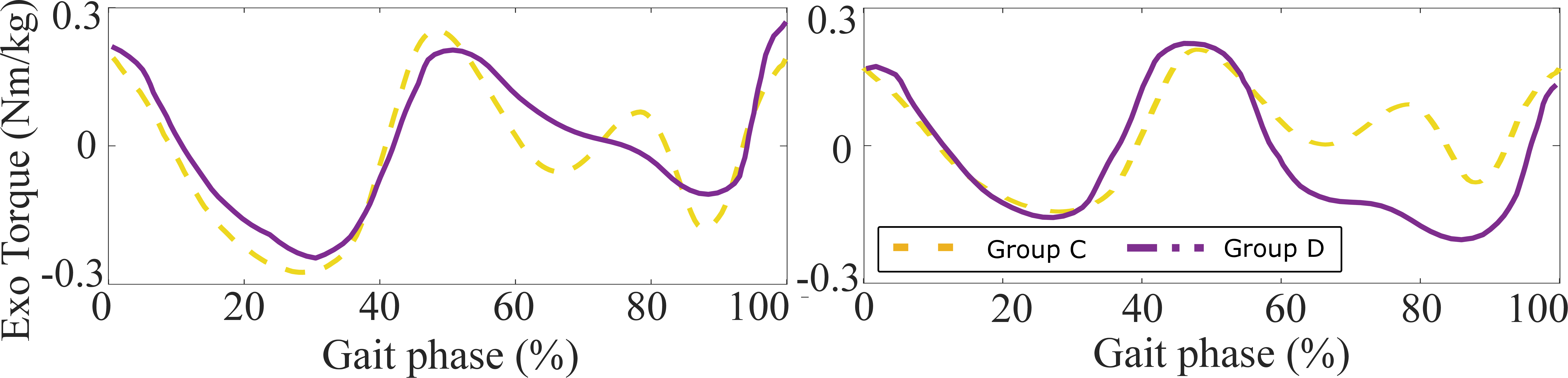}
	\vspace{-1mm}
	\caption{Average exoskeleton torque outputs comparison between Groups C and D on solid ground (left) and sand (right).}
	\label{fig:torques}
\end{figure}

\setcounter{figure}{8}
\begin{figure*}[t!]
	\centering
	\subfigure[]{
		\includegraphics[width = 3.4in]{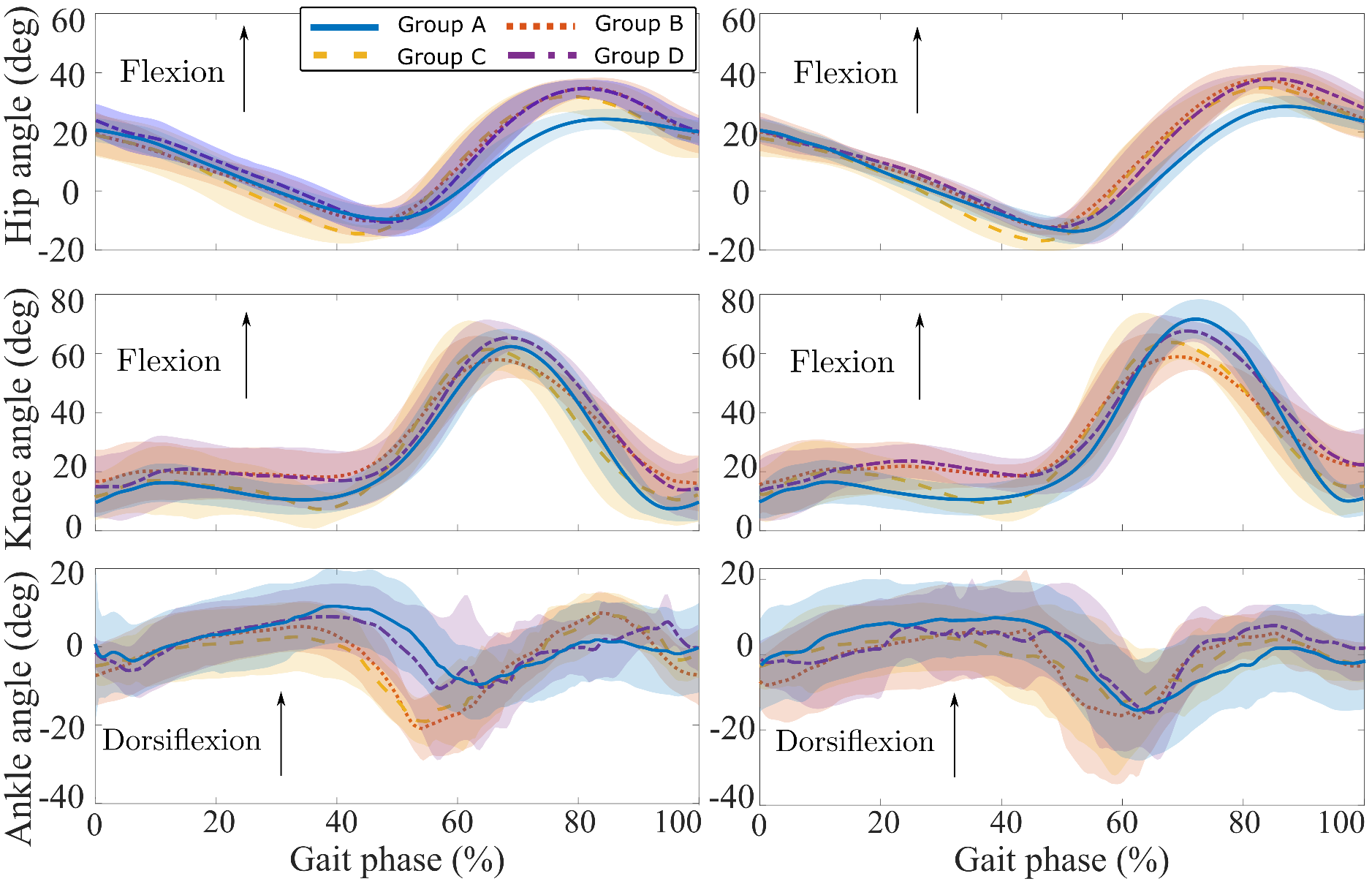}
		\label{fig:angles}}
	\subfigure[]{
		\includegraphics[width = 3.4in]{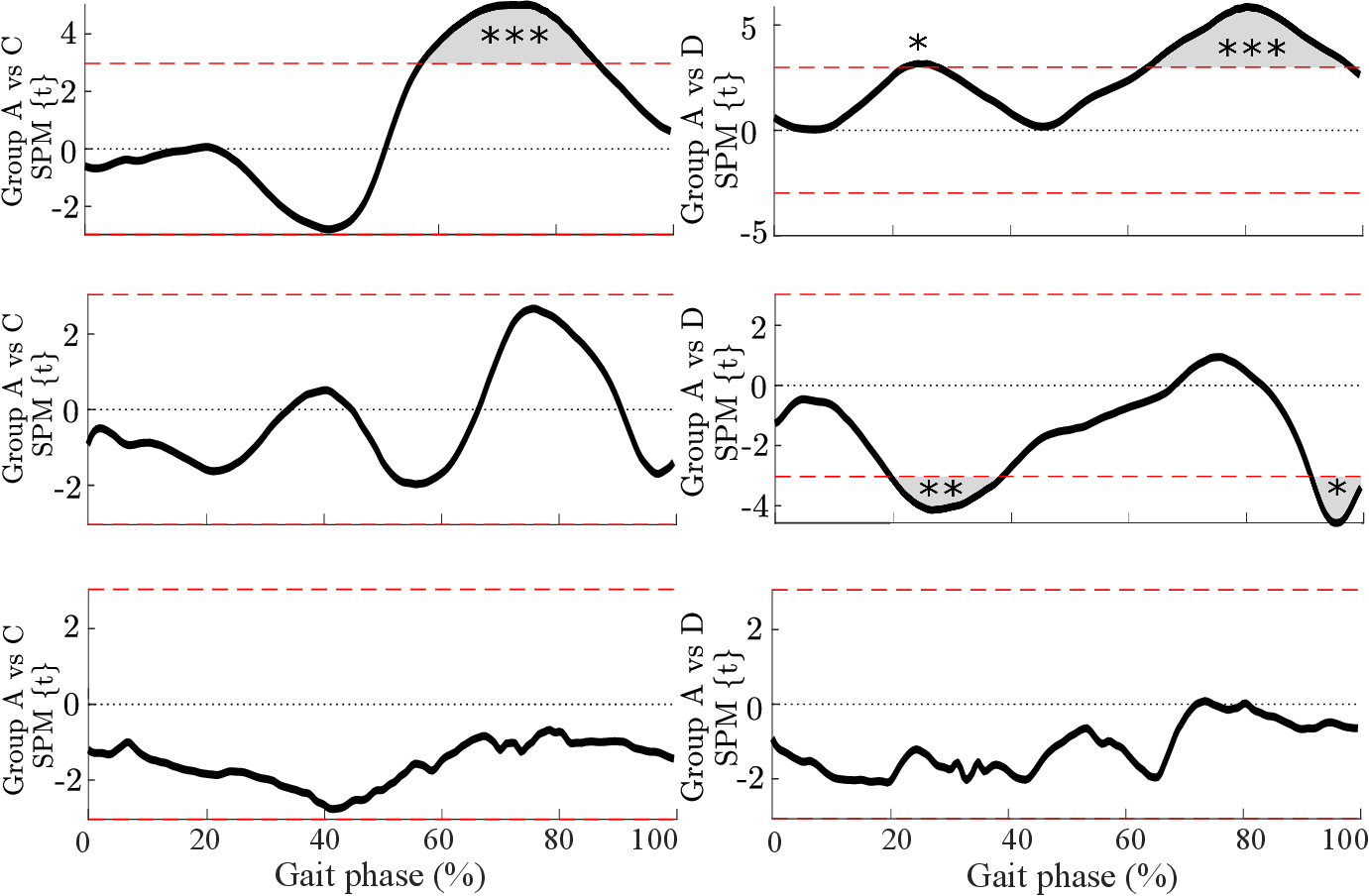}
		\label{fig:SPMangles}}
	\vspace{-2mm}
	\caption{(a) Comparison of hip, knee, and ankle joint angles over one complete gait cycle for solid ground (left column) and sand (right column). The top row: hip flexion(+)/extension(-); the mid row: flexion(+)/extension(-); the bottom row: ankle dorsiflexion(+)/plantarflexion(-). The thick curves are the mean value profiles, while the shaded ares show the one standard deviation around the mean values. (b) The results of 1DSPM statistical analyses comparing walking with the baseline controller (left) vs with the MPC controller (right) on sand. The black curves represent the test statistic (t-value) across the gait cycle, and the red dashed lines indicate the critical threshold for statistical significance at $\alpha = 0.05$. Shaded regions above the threshold indicate phases of the gait cycle where significant differences were detected. Stars ($*$, $**,$ $***)$ denote p-values of $<$0.05, $<$0.01, and $<$0.001, respectively. The x-axis represents the gait phase, normalized from 0 (heel strike) to 1 (subsequent heel strike) for a complete gait cycle.}
	\label{fig:AngleComparison}
\end{figure*}  

\begin{figure*}[t!]
	\centering
	\subfigure[]{
		\includegraphics[width = 3.4in]{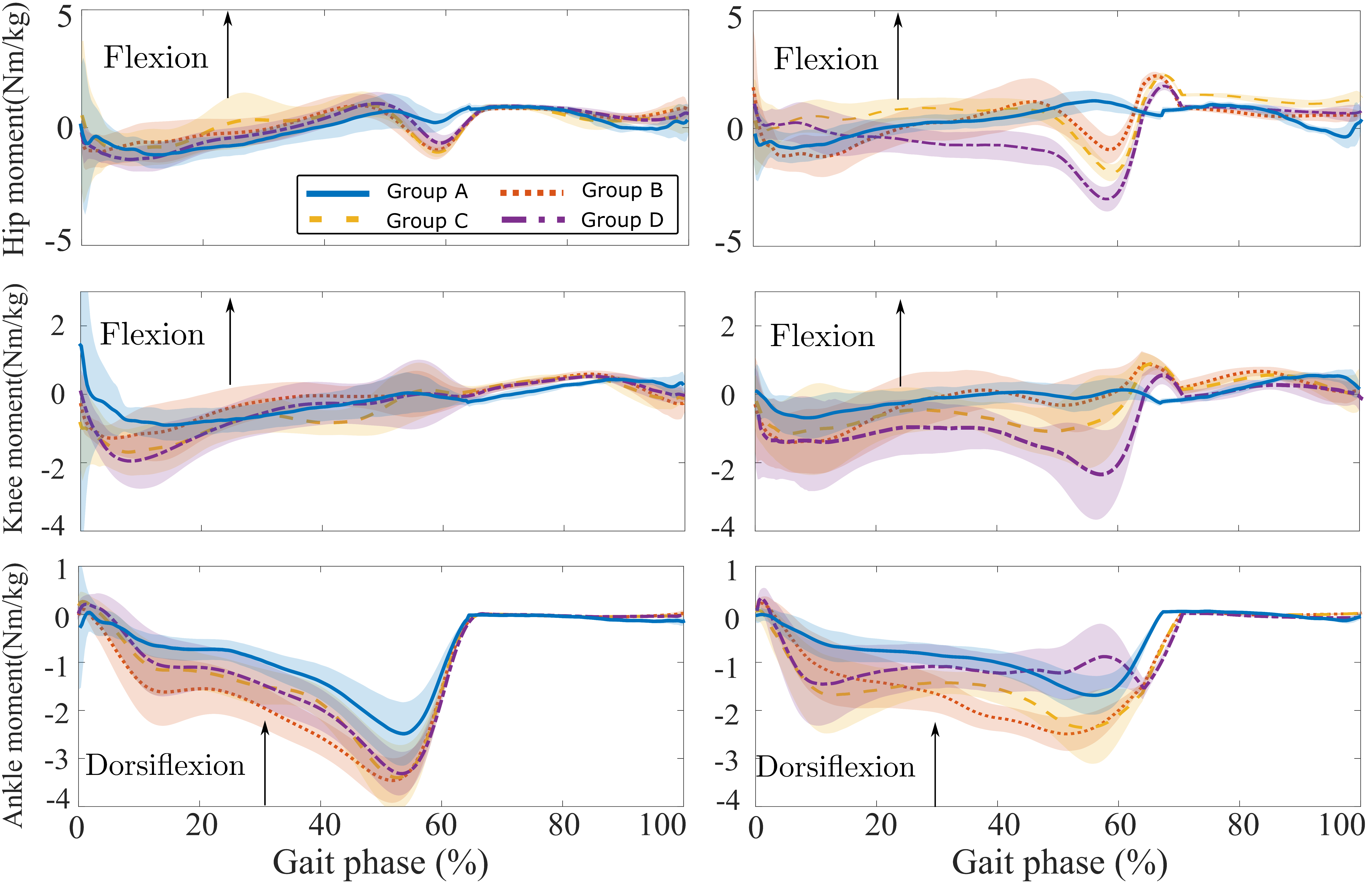}
		\label{fig:moments}}
	\subfigure[]{
		\includegraphics[width = 3.4in]{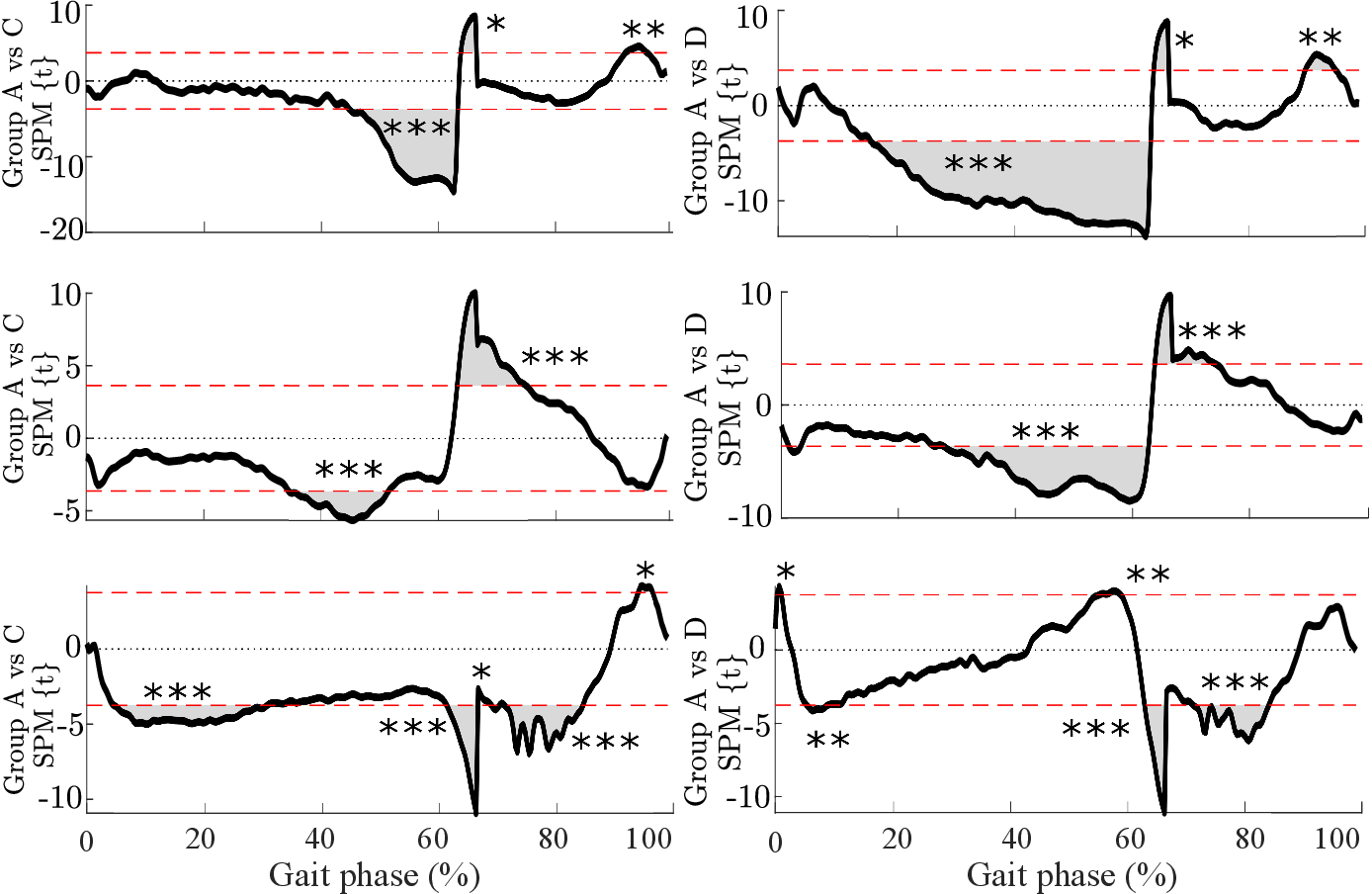}
		\label{fig:SPMmoments}}
	\vspace{-2mm}
	\caption{Comparison of hip, knee, and ankle joint moments over one complete gait cycle for solid ground (left column) and sand (right column). The top row: hip flexion(+)/extension(-); the mid row: flexion(+)/extension(-); the bottom row: ankle dorsiflexion(+)/plantarflexion(-). The thick curves are the mean value profiles, while the shaded ares show the one standard deviation around the mean values. (b) The results of 1DSPM statistical analyses comparing walking with the baseline controller (left) vs with the MPC controller (right) on sand. The black curves represent the test statistic (t-value) across the gait cycle, and the red dashed lines indicate the critical threshold for statistical significance at $\alpha = 0.05$. Shaded regions above the threshold indicate phases of the gait cycle where significant differences were detected. Stars ($*$, $**,$ $***)$ denote p-values of $<$0.05, $<$0.01, and $<$0.001, respectively. The x-axis represents the gait phase, normalized from 0 (heel strike) to 1 (subsequent heel strike) for a complete gait cycle.}
	\label{fig:MomentComparison}
\end{figure*}

Figs.~\ref{fig:GaitOnSolid} and~\ref{fig:GaitOnSand} show the human walking locomotion parameters such as step length, width, and the center of mass (COM) variation of human body on two different terrains. The COM variation is defined as the subject's normalized COM height deviation in the anterior-posterior direction during the gait cycle. It is clear to see an increase in the step length when human subjects wore the knee exoskeletons on the solid ground while the increase became less significant on sand. Paired sample t-tests were conducted to analyze the differences between each group. The MPC-based exoskeletons enabled subjects to walk with the widest step width on sand and the narrowest on solid ground, while maintaining minimal COM variation on both terrains.

Fig.~\ref{fig:angles} shows the hip, knee, and ankle joint angle profiles of four groups on the solid ground and sand. The hip joint angles exhibit greater flexion angles during the swing phase with the exosekelton. The knee kinematics show no significant deviation in the early stance phase (i.e., 0-20\% of the gait phase). A relatively flat knee flexion angle during stance phase (i.e., $0$--$65$\% of gait phase) and a decrease in maximum flexion during swing phase (i.e., $65$--$100$\% of gait phase) on sand were observed. The ankle joint demonstrated greater plantarflexion during mid-stance on sand, particularly in Group~C. We also conducted a statistical analysis across the entire gait cycle using one-dimensional statistical parametric mapping (1DSPM) to capture joint angle variations. The results comparing walking with the baseline controller (left) vs with the MPC controller (right) on sand are presented in Fig.~\ref{fig:SPMangles}. For the hip joint, two main regions of statistical significance ($p<0.001$) were observed within the gait cycle under both controller, i.e., during early to mid swing phase. Under the MPC controller, the knee joint exhibited two significant regions of statistical significance during the mid stance ($p<0.01$) and terminal swing phase ($p<0.05$) of the gait cycle. Regarding the ankle joint, no region of statistical significance throughout the gait is obtained under either controller.  

Fig.~\ref{fig:moments} shows the joint torque profile (normalized by body weight) of the hip, knee, and ankle joints across the gait cycle on solid ground and sand. The 1DSPM analysis results between walking on sand with the baseline controller vs with MPC controller are also presented in Fig.~\ref{fig:SPMmoments}. Although the exoskeleton applied the torques only at the knee joint, the joint torque profiles change significantly at the hip and ankle joints as well. The hip joint displayed an increased extension moment on both terrains with the exoskeleton during early swing phase, with walking on sand showing more significant increases and longer periods, corresponding to the larger hip flexion angles. On solid ground, the knee joint exhibited a characteristic biphasic pattern with peak extension moments during early stance and late stance. On sand, the MPC controller leads to higher knee joint moments during the late stance to early swing phase compared to both the unpowered exoskeleton (Group~B) and the baseline controller (Group~C). The knee moments during late stance and early swing phase was significantly increased across all conditions ($p<$ 0.001). In contrast, the ankle joint moments demonstrated increased variability on sand, particularly during push-off. Group~D showed a significant reduction ($p<$ 0.001) in peak ankle plantarflexion moments compared to other three groups, while the baseline controller increased the plantarflexion moments significantly ($p<$ 0.001).

\begin{figure*}[h!]
	\centering
	\vspace{-2mm}
	\subfigure[]{
		\includegraphics[width = 3.2in]{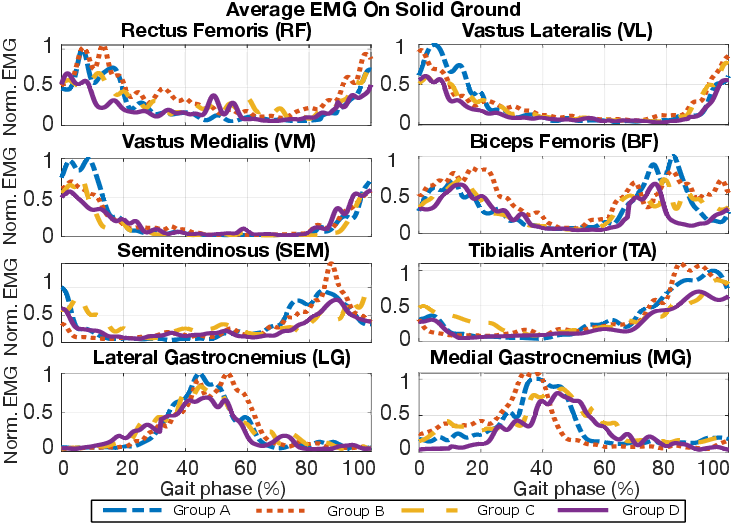}
		\label{fig:meanEMG_solidground}}
	\hspace{-1mm}
	\subfigure[]{
		\includegraphics[width = 3.2in]{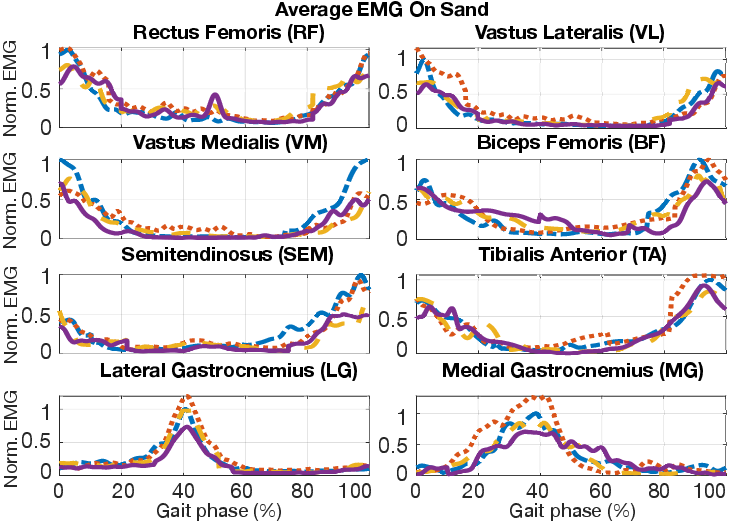}
		\label{fig:meanEMG_sand}}
	\label{fig:meanEMG}
	\vspace{-2mm}
	\caption{Normalized EMG results for $8$ muscles of a randomly selected subject on (a) solid ground and (b) sand. }
\end{figure*}  

\begin{figure*}
	\hspace{-3mm}
	\subfigure[]{
		\includegraphics[width = 7.1in]{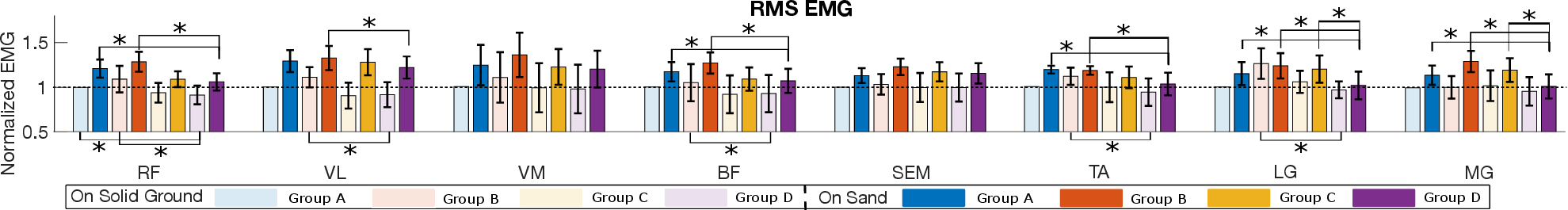}
		\label{fig:EMG_RMS}}
	\hspace{-4mm}
	\subfigure[]{
		\hspace{-3mm}
		\includegraphics[width = 7.1in]{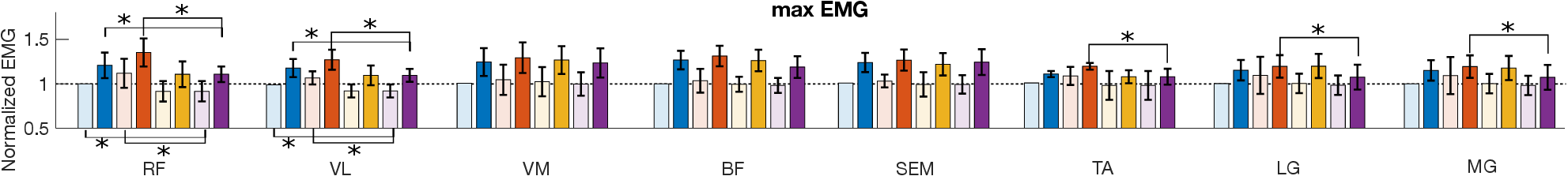}
		\label{fig:EMG_max}}
	\label{fig:EMG}
	\caption{Normalized average (a) RMS and (b) maximum EMG activation under four conditions for all subjects on both terrains. Statistical analysis was conducted between Group D and the other groups on each terrain. The label $*$ indicates a significant ($p < 0.05$) difference between groups.}
\end{figure*}

\subsection{Muscle Activation and Metabolic Cost}
Figs.~\ref{fig:meanEMG_solidground} and~\ref{fig:meanEMG_sand} show the EMG signals for eight muscles from a representative subject, normalized with respect to those of Group A on solid ground. Both RMS and maximum EMG metrics are included because RMS provides a measure of the overall muscle activity over time, while maximum EMG captures the peak activation. Peak EMG signals were highest in the unpowered condition (Group B) for both terrains, attributed to increased mass and resistance from the exoskeleton. On solid ground, both controllers showed comparable muscle activation reductions. However, the MPC led to notable decreases in peak EMG for ankle extensors and flexors (TA, LG, and MG), particularly on sand. Walking on granular terrain requires coordinated control across the lower limbs. Although torque was applied at the knee, the reduced muscle activation at the ankle shows that the exoskeleton assistance had a positive effect on the dynamics of the entire lower limb.

The RMS and maximum EMG values have been normalized to those of Group~A on solid ground, accounting for all trials and subjects across the four groups. Figs.~\ref{fig:EMG_RMS} and \ref{fig:EMG_max} show the results of this analysis. Among all trials, Group B consistently exhibited the highest activity on both terrains. Statistical analysis revealed significant decreases in RF, VL, LG, and MG muscles for Group D on both terrains. Table~\ref{tbl:EMG_summarize} summarizes the RMS and maximum muscle activity percentage changes for Groups~C and D relative to Group~A, where positive values indicate an increase in muscle activity and negative values represent a reduction. Both controllers demonstrated similar trends in alleviating muscle activation across terrains. On solid ground, Group~C showed slight advantages in VL and BF RMS values, while Group~D significantly reduced LG and MG activation. On sand terrain, Group~D showed pronounced reductions, particularly in RF, TA, LG, and MG RMS values. Although the SEM activation increased on sand for both groups, Group D showed smaller increases in both RMS and maximum values.

{\renewcommand{\arraystretch}{1.2}
	\setlength{\tabcolsep}{0.05in}
	\begin{table}[h!]
		\centering
		\caption{RMS/Max EMG comparison in percentage change (\%)} 
		\begin{tabular}{c|cc|cc}
			\hline
			\multirow{2}{*}{Muscle} & \multicolumn{2}{c}{Groups C vs. A}  & \multicolumn{2}{c}{Groups D vs. A} \\ \cline{2-5} 
			&  Solid  & Sand  & Solid & Sand \\ \hline
			RF  & $-6.0/-8.4$  & $-9.8/-8.4$ & $-8.6/-8.4$ & $-12.3/-8.3$  \\ \hline
			VL  & $-9.6/-7.0$ & $-0.9/-6.9$ & $-8.5/-6.9$   & $-3.0/-6.9$   \\ \hline
			VM  & $-1.4/1.8$   & $-1.6/-1.7$ & $-2.7/-0.84$  & $-3.7/-0.84$   \\ \hline
			BF  & $-8.2/-0.4$  & $-7.0/-0.39$ & $-7.4/-1.7$  & $-8.7/-1.7$  \\ \hline
			SEM & $-0.30/-1.6$ & $3.8/-1.6$ & $-0.33/-1.6$   & $\boldsymbol{2.2}/\boldsymbol{1.6}$     \\ \hline
			TA  & $-0.71/-2.7$ & $-12.6/-2.7$ & $-6.2/-2.7$  & $-13.5/-2.7$  \\ \hline
			LG  & $\boldsymbol{5.79}/\boldsymbol{0.21}$    & $\boldsymbol{4.4}/\boldsymbol{4.8}$  & $-3.2/-1.8$   & $-11.5/-6.7$  \\ \hline
			MG  & $\boldsymbol{2.29}/\boldsymbol{0.25}$    & $\boldsymbol{5.1}/\boldsymbol{4.7}$ & $-4.0/-2.8$    & $-8.7/-7.6$   \\ 
			\hline
		\end{tabular}
		\label{tbl:EMG_summarize}
	\end{table}
}

\begin{figure}[h!]
	\centering
	\includegraphics[width = 3in]{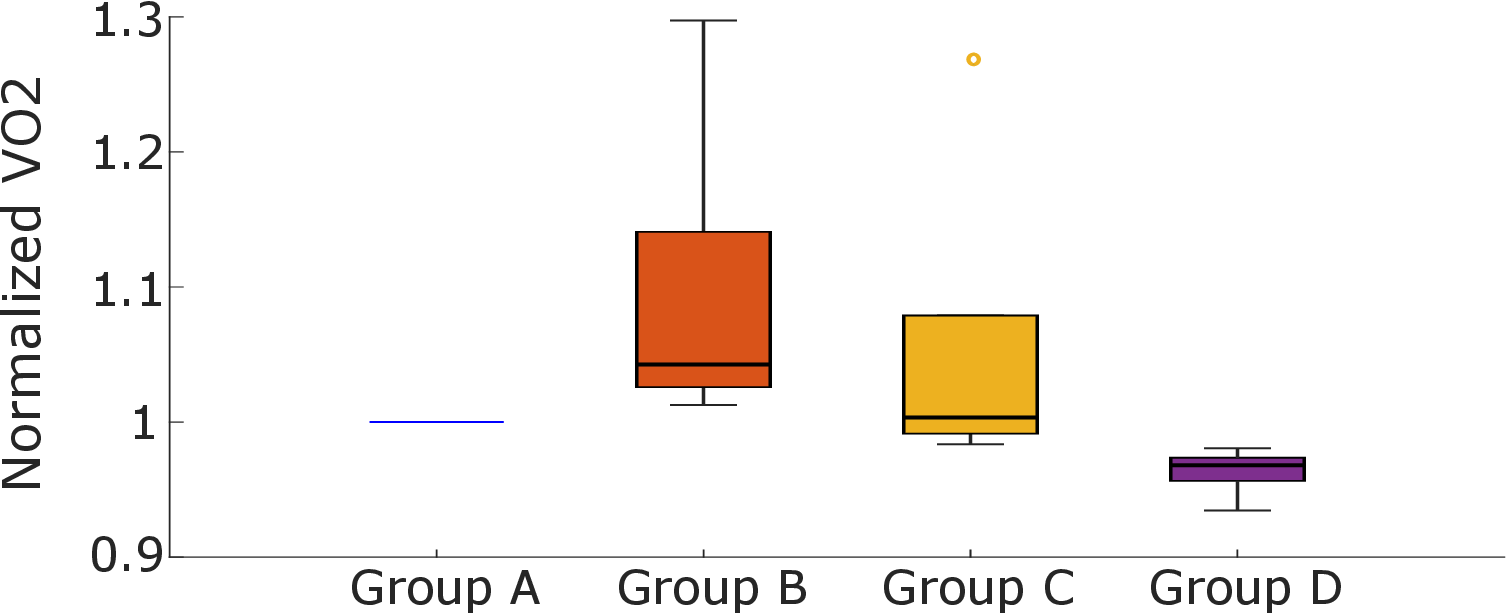}
	\caption{Normalized VO2 results ofoutdoor experiments. The results of Group B, C, and D are normalized with respect to Group A.}
	\label{fig:VO2Results}
	\vspace{-3mm}
\end{figure}

Fig.~\ref{fig:VO2Results} shows the energy expenditure using the normalized VO2 max data. The findings indicate a notable increase in metabolic cost when ambulating on sand with the unpowered exoskeleton. While the baseline controller, optimized for solid ground locomotion, did reduce the metabolic burden, the exertion remained higher compared to no exoskeleton condition. Implementing the MPC controller resulted in a $3.7$\% decrease in metabolic rate relative to without wearing the exoskeleton, underscoring the efficiency of the control strategy in reducing energy cost. Paired sample t-tests revealed no statistically significant differences in metabolic cost between conditions. 

\section{Discussions}
\label{sec:dis}

Walking on sand introduces significant biomechanical challenges, as reflected in the altered joint angle and moment profiles. Analysis of knee joint angles on sand (Fig.~\ref{fig:AngleComparison}) revealed that wearing the exoskeleton constrained the subject's knee joint movement to some degrees. For instance, the unpowered and MPC-controlled modes made less flexion within the certain stance phase (i.e., $25$--$40$\% of gait phase), and for Groups~B, C and D, the maximum knee flexion of swing phase decreased compared with natural walking without exoskeleton. However, the MPC allowed for a gait more closely mimicking unassisted walking compared to other conditions. The improvement in assistance performance was evident in both the angle magnitude and gait phase. The observed differences in ankle plantarflexion and hip flexion angles further emphasize the role of multi-joint coordination in navigating granular terrains. Similarly, joint moment analysis revealed a redistribution of mechanical effort across the lower limbs, especially on sand. The knee joint exhibited significantly increased extension moments throughout the stance phase under the MPC controller, highlighting its compensatory role in maintaining stability and propulsion. This redistribution aligns with the control intent of reducing reliance on hip and ankle propulsion to compensate for the COM sinkage when walking on sand by augmenting knee joint contribution.


The knee joint generates less mechanical power than the hip and ankle but plays a key role in gait stabilization, particularly on sand. Fig.~\ref{fig:angles} shows a flatter knee flexion profile during the stance phase on sand, indicating the knee's role in maintaining stability. By reducing the effort at the knee during the stance phase, the exoskeleton helps distribute forces evenly across the lower limbs, improving walking efficiency. Although the knee contributes minimally to propulsion, the observed reduction in muscle activation suggests the exoskeleton effectively aids in maintaining stability. Supporting knee flexion and extension reduces the need for compensatory strategies that would otherwise heighten ankle stabilization demands, as evidenced by the significant reduction in TA, LG, and MG muscle activations in Group~D, shown in Fig.~\ref{fig:meanEMG_solidground} and summarized in Table~\ref{tbl:EMG_summarize}. This, in turn, explains the observed decrease in TA muscle activity during the swing phase, highlighting how knee assistance indirectly benefits the entire lower limb and contributes to improved gait efficiency. The results demonstrate that the MPC consistently outperforms other conditions in reducing muscle activation and improving stability on sand, as evidenced by the notable reductions in muscle activations in Table~\ref{tbl:EMG_summarize}, supporting the main claim that the MPC design offers more effective assistance than standard methods.

The observed redistribution of joint moments highlights a trade-off between reducing muscle activation and managing energy expenditure. By redistributing mechanical efforts among the lower limbs, the exoskeleton reduces localized muscle fatigue. Specifically, the MPC controller significantly reduced activation in several lower-limb muscles on both terrains as summarized in~Table~\ref{tbl:EMG_summarize}, including ankle extensors and flexors (TA, LG, and MG). This aligns with findings in~\cite{Zhu2024SandJBME} regarding increased work needed on sandy terrains to maintain COM height and forward momentum. The increase in SEM muscle activation on sand for both Groups~C and D highlights the challenging nature of sandy terrain, which requires additional muscle engagement for stability but not fully offset by the exoskeletons. In contrast, the unpowered exoskeleton (Group~B) increased muscle activation due to added weight and lack of active control, reinforcing the importance of adaptive torque assistance. These observations imply that refined exoskeleton design and control strategies can accommodate the complex walking dynamics on sand.

Walking on sand requires $1.6$--$2.5$ times more mechanical work and metabolic cost than on solid ground~\cite{lejeune1998mechanics}. Unpowered exoskeletons exacerbated high energy cost, increasing muscle activation and metabolic cost (see Figs.~\ref{fig:EMG_RMS} and~\ref{fig:VO2Results}). It was reported that knee exoskeleton did not provide a benefit in the cases where small positive mechanical work was required at the knee joint such as walking on flat surfaces because of the additional mass of the exoskeleton itself~\cite{maclean2019energetics}. The unpowered mode and baseline controller increased metabolic cost by 9.4\% and 5.4\%, respectively, compared to unassisted walking, aligning with previous findings of 8-15\% increases on uneven terrain~\cite{kowalsky2021human}. This increase is correlated with additional effort required to stabilize the body on sand. The unpowered exoskeleton not only adds weight but also restricts natural movement, increasing the metabolic burden. Similarly, while the baseline controller offers some assistance, it still demands extra effort to maintain balance and propulsion, which offsets the potential benefits of reduced knee muscle activation. In contrast, the MPC controller decreased metabolic cost by 3.7\% comparing to the normal walking on sand without exoskeleton. Though, no statistical significance was found on the metabolic cost. One possible reason for this lack of significance could be differences in the physical fitness of subjects, with some participants potentially having strong endurance or conditioning, which may need further investigation.

The above findings have significant implications for rehabilitation and assistive device design. The ability of the proposed MPC to enhance gait stability and reduce muscle activation on sand suggests potential applications for individuals with lower-limb impairments, particularly in environments with variable terrain compliance. By redistributing mechanical effort and stabilizing joint movements, the exoskeleton can improve user's mobility and reduce muscle fatigue. On the other hand, a discrepancy was observed between muscle activation reduction and metabolic cost. For instance, the unpowered exoskeleton (Group~B) significantly increased metabolic cost by 9.4\%. While the baseline controller (Group~C) reduced activation in the RF, VM, BF, and TA muscles on sand, the overall metabolic cost increased. These findings may be due to the extra effort required to carry the exoskeleton and stabilize other joints, which offset the benefits of knee assistance. This underscores the need for further investigation into the relationship between muscle synergy and metabolic cost, as well as the optimization of assistive device design and control strategies, including assistance level ($\alpha$) and MPC parameters. Incorporating multi-joint coordination into the control framework could further enhance stability and reduce metabolic cost by distributing effort more effectively across the lower limbs. Additionally, the current controller does not explicitly account for speed variations, and complex foot-sand interactions contribute to GRF fluctuations. The experiment setup and design are further needed for accurate GRF measurements. We plan to explore the interplay between spatial and temporal features to develop adaptive control strategies for various walking conditions, including different terrains and speeds, ultimately enhancing the system's robustness and real-world applicability.

\section{Conclusions}
\label{sec:con}

This paper presented an integrated knee exoskeleton control for augmenting human locomotion on solid ground and sand. The machine learning-based GRF estimation enabled MPC for stiffness-based efficiently walking gait control on sand. The experimental findings validated the effectiveness of the proposed MPC controller in improving walking gait dynamics in both muscle activation reductions and less metabolic cost. The results also demonstrated that knee exoskeletons can help redistribute the kinematic and kinetic changes among the lower-limb joints. These insights underscore the complexity of human-exoskeleton interaction on granular terrains and highlight the need for a holistic understanding of human locomotion biomechanics and energetic trade-offs. Future work includes optimizing the exoskeleton design and enhancing the control strategy to provide comprehensive assistance.

\bibliography{Ref_2024JDSMC}

\end{document}